\documentclass{article}
\usepackage[preprint]{log_2025}			

\usepackage{booktabs}						
\usepackage{multirow}						
\usepackage{amsfonts}						
\usepackage{graphicx}						
\usepackage{duckuments}						
\usepackage{subcaption} 
\usepackage[numbers,compress,sort]{natbib}	

\title[Graph Contrastive Learning vs Untrained Baselines: The Role of Dataset Size]{Graph Contrastive Learning versus Untrained Baselines: The Role of Dataset Size}

\author[Khanna et al.]{Smayan Khanna, Doruk Efe Gökmen, Risi Kondor, Vincenzo Vitelli \\
University of Chicago \\
\texttt{\{smayan, gokmen, risi, vitelli\}@uchicago.edu}}

\begin{document}

\maketitle

\begin{abstract}
Graph Contrastive Learning (GCL) has emerged as a leading paradigm for self-
supervised learning on graphs, with strong performance reported on standardized
datasets and growing applications ranging from genomics to drug discovery. We ask a basic question: \emph{does GCL actually outperform untrained baselines?} We find that GCL's advantage depends strongly on dataset size and task difficulty. On standard datasets, untrained Graph Neural Networks (GNNs), simple multilayer perceptrons, and even handcrafted statistics can rival or exceed GCL. On the large molecular dataset \textsc{ogbg-molhiv}, we observe a crossover: GCL lags at small scales but pulls ahead beyond a few thousand graphs, though this gain eventually plateaus. On synthetic datasets, GCL accuracy approximately scales with the logarithm of the number of graphs and its performance gap (compared with untrained GNNs) varies with respect to task complexity. Moving forward, it is crucial to identify the role of dataset size in benchmarks and applications, as well as to design GCL algorithms that avoid performance plateaus.



 
\end{abstract}

\section{Introduction} \label{introduction}

Graph Contrastive Learning (GCL) has emerged as a prominent approach within self-supervised learning, where models learn graph and node representations by contrasting positive and negative views derived from the data itself \cite{ju2024gclreview}. This approach has been widely studied, with strong performance reported on standard benchmarks and growing interest in applications such as drug discovery \cite{dd_1, dd_2}, genomics \cite{gen_1, gen_2}, and social-network analysis \cite{social_1,social_2}. 
Yet it remains unclear \emph{when} GCL truly offers an advantage. We find that simple baselines such as untrained GNNs, MolFingerprint MLPs \cite{errica_2020}, and handcrafted graph statistics can rival GCL on small TU datasets \cite{tu_datasets}, which remain widely used by recent GCL papers \cite{topoGCL,reinforce24,MSSGCL,CTaug,khanGCL} despite their limited size (e.g., \textsc{MUTAG}: 188 graphs; \textsc{PROTEINS}: 1,113).

This motivates us to analyze how dataset size influences GCL’s advantage relative to untrained baselines for graph-classification tasks. By subsampling \textsc{ogbg-molhiv} \cite{ogb, ogb_molhiv_paper}, a large dataset of HIV-inhibiting molecules, we observe a crossover at roughly 4k graphs where GCL begins to pull ahead, though gains eventually plateau. As a controlled check, we use synthetic datasets with adjustable size and difficulty \cite{trivedi_two} to show how task complexity shapes these scaling effects. 

\textbf{Related Work.} \label{related_work}
 Prior work has argued for including untrained GNNs in evaluation for both supervised graph learning and GCL \cite{errica_2020,trivedi_one}, with limited adoption in practice. Handcrafted heuristics have been used to benchmark other graph learning settings, such as link-prediction, \cite{heuristic_cai_22,gurukar2022benchmarking}, but not for \emph{graph-level GCL}. Trivedi et al. \cite{trivedi_two} analyzed the gap between untrained baselines and GCL in terms of invariances and released a synthetic dataset generator which we adopt in our work. Distinct from their focus, we analyze this gap under the lens of dataset size. See App.~\ref{rel_work} for a more detailed discussion. 

\section{Benchmarking GCL with Simple Baselines} \label{benchmarking_simple}
We focus on four representative GCL methods: GraphCL, JOAO, SimGRACE, and InfoGraph \cite{gclyu, joao, simGRACE22, Sun2020InfoGraph} (see App. \ref{background}). 
We compare them with simple baselines under the standard embedding–probe protocol \cite{Sun2020InfoGraph}, which evaluates representation quality by freezing the pretrained encoder and training a lightweight linear classifier on top of embeddings. Results are averaged over 5 random seeds, and we quote prior numbers where experimental settings match. See App.~\ref{eval_protocol} for a full table of dataset statistics and more information on the evaluation protocol.

\textbf{Simple baselines.} We include: 
(i) \textbf{Untrained GNN}: a 3-layer GIN~\cite{xu2018how} with sum pooling, matching the exact encoder architecture used by GCL papers \cite{Sun2020InfoGraph}. 
(ii) \textbf{Handcrafted Graph statistics}: a fixed-dimensional descriptor per graph comprising node count, mean degree, and a 5-bin degree histogram; features are normalized within each training fold before SVM. 
(iii) \textbf{MolFingerprint}: an untrained 256-dimensional MLP with global pooling following Errica et al.~\cite{errica_2020}; applicable only to molecular datasets with node features. We also include other baselines such as graph kernels, with a full table of results and ablations reported in App.~\ref{benchmark_res}.

\begin{figure}[t]
  \centering
  \includegraphics[width=\linewidth]{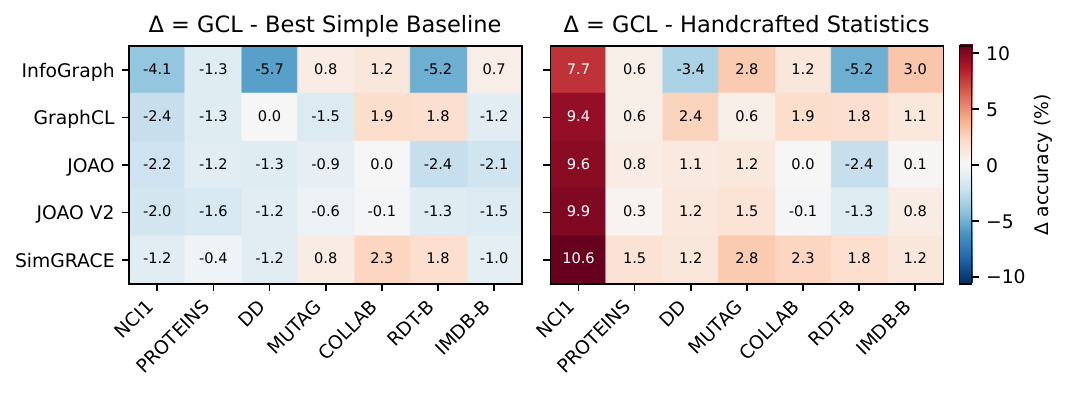}
  \caption{Graph classification accuracy difference (\%) between GCL and baselines. Rows show representative GCL methods, columns the TU datasets. Left: GCL vs the best-performing simple baseline for that dataset. Right: GCL vs.\ the handcrafted statistics baseline. Negative values (blue) indicate that the baseline outperforms the GCL method, while positive values (red) favor GCL. 
 }
  \label{fig:tu_heatmaps}
\end{figure}

\textbf{Findings.}
Across seven TU datasets, simple baselines prove surprisingly competitive. On molecular datasets with node features, untrained GNNs and MolFingerprint often match GCL. Among the baselines we test, fixed handcrafted statistics, a very minimal descriptor, perform best overall, closely tracking GCL on most tasks and occasionally outperforming it (Fig.~\ref{fig:tu_heatmaps}, right panel). In fact, every GCL method is beaten on at least some datasets by an untrained or simple alternative (Fig.~\ref{fig:tu_heatmaps}, left panel). These findings suggest that part of GCL’s reported advantage on TU benchmarks may stem from dataset-specific effects \cite{rethinking_graphclass_li} rather than intrinsic benefits of contrastive learning, motivating our scaling experiments.

\section{Scalability case study: \textsc{ogbg-molhiv}} \label{molhiv_study}

We extend our analysis to \textsc{ogbg-molhiv} \cite{ogb, ogb_molhiv_paper}, a molecular prediction dataset with 41{,}127 graphs, which is an order of magnitude larger than the TU datasets we considered in Sec. \ref{benchmarking_simple}. For computational feasibility we focus on a single representative method, GraphCL, chosen for its wide adoption, competitive reported accuracy, and relative simplicity \cite{gclyu}. Using OGB’s official scaffold split, we pretrain GraphCL only on the training partition and evaluate the resulting encoder on train/validation/test embeddings following \cite{PyGCL21}. A logistic regression probe is trained on train embeddings, tuned on validation, and evaluated on test ROC-AUC to account for class imbalance. The encoder backbone is a 3-layer edge-aware GINE \cite{GINE}, and its randomly initialized, frozen version serves as our untrained baseline. We also include the handcrafted graph statistics baseline from Sec. \ref{benchmarking_simple}.

\textbf{Scaling setup.} We subsample the training set at fractions $\{1,2,5,10,12.5,15,20,30,50,75,100\}\%$ using 5 random seeds. For each fraction $f$, GraphCL is pretrained on the $f$-subset for 100 epochs with edge-dropping augmentations, with hyperparameters tuned on a small grid search. The linear probe is trained on embeddings of the same $f$-subset, while validation and test embeddings are always computed on the full splits. Both baselines we test follow the identical pipeline and use the same $f$-subset for the probe, ensuring a fair comparison. We report mean $\pm$ standard deviation across the 5 seeds. See App. \ref{hpo} for more on the evaluation protocol, training setup and hyperparameter search.

\begin{figure}[t]
  \centering
  \includegraphics[width=\linewidth]{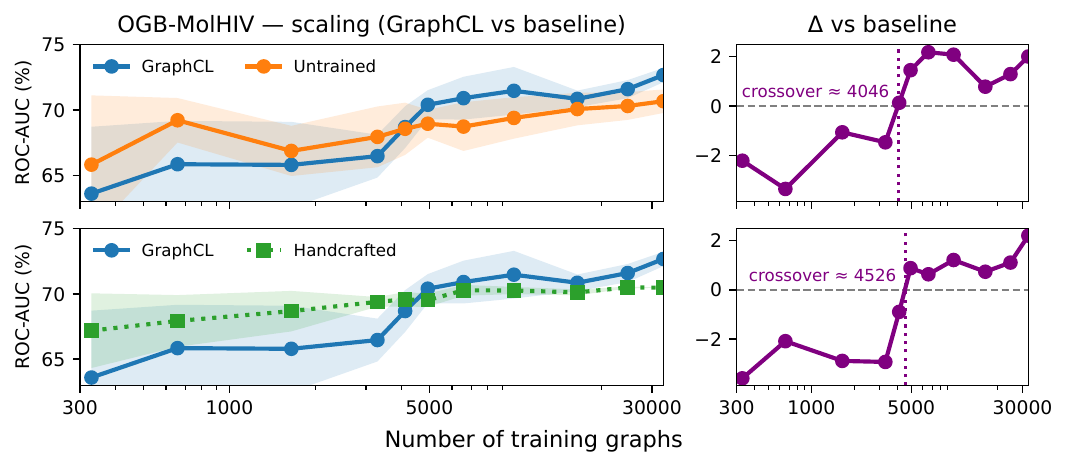}
  \caption{\textbf{Scaling on \textsc{ogbg-molhiv} (Log Scale).} \emph{Top:} GraphCL vs.\ an untrained GINE. \emph{Bottom:} GraphCL vs.\ the handcrafted statistics baseline. In each row, the left panel shows test ROC-AUC vs.\ number of training graphs with shaded $\pm 1\sigma$ over seeds; the right panel shows $\Delta$ (GraphCL$-$baseline, in percentage points) with the vertical dotted line marking the approximate crossover where $\Delta=0$. Positive values favor GraphCL.}

  \label{fig:ogbg_scale}
\end{figure}

\textbf{Results.} Figure~\ref{fig:ogbg_scale} shows a crossover behavior for both baselines.
At ($\leq{\sim}4{,}000$ training graphs), both the untrained GINE and the handcrafted statistics baseline can match or even outperform GraphCL. 
Beyond ${\sim}4.5$k training graphs, GraphCL overtakes both baselines, but its advantage remains modest: the gap stabilizes at about ${\sim}2$\% ROC-AUC up to the largest scale (${\sim}33$k). This suggests that scale is \emph{necessary} for contrastive gains to appear, but that adding more data alone does not widen the gap further. Importantly, the crossover region for both baselines is not trivially small: it sits at or above the size of most TU graph-classification datasets we benchmark in Sec.~\ref{benchmarking_simple}. We also note that variance is highest in the low-data regime, where subsampling and augmentation randomness have greater effect, and decreases at larger scales. Taken together, these results highlight that dataset size is a critical axis for GCL evaluation, even though the exact advantage threshold is dataset and setup-dependent.

\section{Synthetic dataset study}
 We use Trivedi et al.'s generator \cite{trivedi_two} to construct controllable synthetic graph-classification datasets. We stress that these datasets are not intended to serve as a definitive method-ranking benchmark but rather provide a controlled sandbox to probe how GCL behaves. Each graph consists of a label-defining \emph{motif} graph attached to a label-agnostic random \emph{background} tree graph. A multiplier $S$ scales the background and thus controls task difficulty, while total dataset size is varied independently. We instantiate six distinct motif graphs (corresponding to six classes), sweep $S\in\{2,4,6\}$, and vary dataset size over $\{600, 3{,}000, 6{,}000, 12{,}000\}$ graphs. Classes are balanced. 

\textbf{Training setup.} We test two representative GCL methods: GraphCL (with node- and edge-dropping augmentations) and InfoGraph. Both are trained for 50 epochs at each dataset size, with hyperparameters and architectures detailed in App.~\ref{synth_info}. As baselines, we include the corresponding untrained GNN encoder and the handcrafted graph statistics features from Sec.~\ref{benchmarking_simple}. Evaluation follows the embedding–probe protocol of App.~\ref{eval_protocol}, and we report classification accuracy averaged over 5 seeds.

\begin{figure}[t]
  \centering

  \begin{subfigure}{0.92\linewidth} 
    \centering 
    \includegraphics[width=\linewidth]{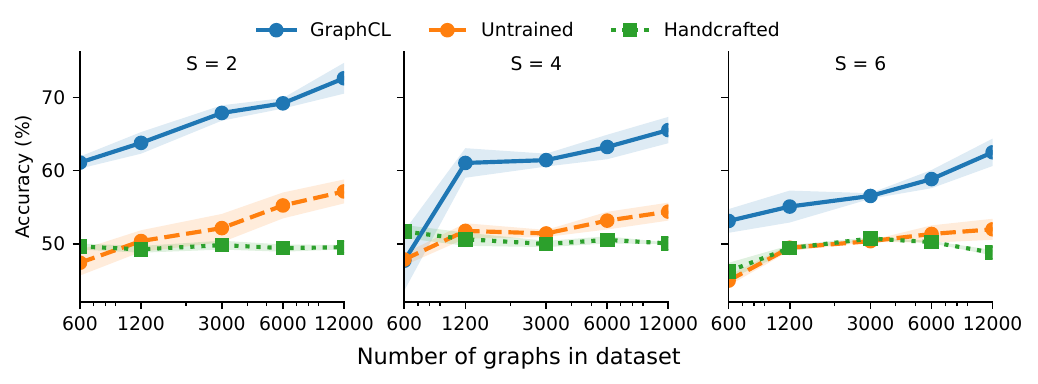}
  \end{subfigure}

  \begin{subfigure}{0.92\linewidth}
    \centering
    \includegraphics[width=\linewidth]{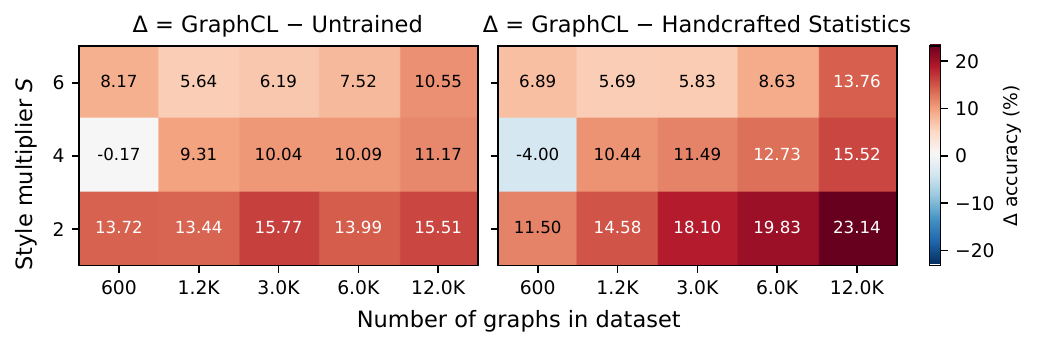}
  \end{subfigure}

  \caption{\textbf{Scaling on Synthetic Datasets.}
    \emph{Top:} GraphCL. Test accuracy vs.\ total number of graphs (log scale) for $S\in\{2,4,6\}$; bands show $\pm1\sigma$ over seeds; handcrafted baseline shown as dotted line.
  \emph{Bottom:} $\Delta$ heatmaps for GraphCL—Untrained and GraphCL—Handcrafted Statistics baseline across style multiplier $S$ and dataset size (shared color scale; positive values favor GraphCL).}
  \label{fig:graphcl_composite}
\end{figure}

\textbf{Results.} Figure~\ref{fig:graphcl_composite} highlights how scale and difficulty shape GCL’s advantage. At $S{=}2$ (easy), GraphCL is consistently better but the gap grows only slowly, showing that extra data helps both it and the untrained encoder. At $S{=}4$ we see a low-data crossover where baselines win at 600 graphs but GraphCL overtakes once more examples are available. At $S{=}6$ (hardest), GraphCL dominates throughout. InfoGraph shows similar trends across dataset size and difficulty (App.~\ref{synth_info}, Fig. \ref{fig:infograph_composite}). Interestingly, it outperforms GraphCL on this generator, consistent with prior findings that generic augmentations can obscure motif signal \cite{trivedi_two}. Being augmentation-free, InfoGraph avoids this pitfall.

A notable pattern is that both GCL and the untrained encoder continue to benefit from more data, growing approximately logarithmically with dataset size, while handcrafted features plateau early, likely because fixed statistics cannot capture new signal (due to limited variability in generated synthetic graphs). It suggests that untrained GNNs are non-trivial baselines at scale: they act like high-capacity random feature maps that improve as the probe sees more labels. GCL’s gains reflect an added benefit from contrastive pretraining, and the gap is clearer here than on \textsc{ogbg-molhiv} likely because we instantiate constant features for \emph{all} nodes in the synthetic graphs (see App. \ref{synth_info}), so the untrained baseline is weaker than in molecular graphs with rich attributes. Additionally, labels in the synthetic dataset are perfectly determined by motifs, so fixed statistics are less informative, whereas in TU datasets label signal has been shown to correlate with graph statistics \cite{rethinking_graphclass_li}, explaining why the handcrafted baseline performs better there.






\vspace{-2mm}



\section{Conclusion}

We showed that: (i) untrained baselines often rival GCL on TU datasets, (ii) on \textsc{ogbg-molhiv} both GCL and an untrained encoder improve with scale before GCL pulls ahead and plateaus, and (iii) on a synthetic suite both grow while handcrafted features saturate. Taken together, these results offer two practical takeaways. First, simple baselines provide useful diagnostics, clarifying when GCL’s gains are substantive versus marginal. Second, for both benchmarks and applications, examining performance as a function of scale relative to such baselines is informative, since non-trivial crossover points can emerge, as we observed. Looking forward, a key challenge is to better quantify this scaling behavior. While our results suggest approximately logarithmic improvements in predictive performance with dataset size, they need to be corroborated by larger datasets. Questions remain about why crossovers occur, when gains saturate, and what governs their magnitude. This points to the need for scaling laws that generalize across datasets, architectures, and tasks. Addressing these questions will likely require not just larger benchmarks, but also methods designed to sustain improvements at scale.

\newpage
\bibliographystyle{unsrtnat}
\bibliography{reference}

\begin{thebibliography}{38}
\providecommand{\natexlab}[1]{#1}
\providecommand{\url}[1]{\texttt{#1}}
\expandafter\ifx\csname urlstyle\endcsname\relax
  \providecommand{\doi}[1]{doi: #1}\else
  \providecommand{\doi}{doi: \begingroup \urlstyle{rm}\Url}\fi

\bibitem[Ju et~al.(2024)Ju, Wang, Qin, Mao, Xiao, Luo, Yang, Gu, Wang, Long, Yi, Luo, and Zhang]{ju2024gclreview}
Wei Ju, Yifan Wang, Yifang Qin, Zhengyang Mao, Zhiping Xiao, Junyu Luo, Junwei Yang, Yiyang Gu, Dongjie Wang, Qingqing Long, Siyu Yi, Xiao Luo, and Ming Zhang.
\newblock Towards graph contrastive learning: A survey and beyond, 2024.

\bibitem[Tao et~al.(2023)Tao, Liu, Lin, Song, and Zeng]{dd_1}
Wen Tao, Yuansheng Liu, Xuan Lin, Bosheng Song, and Xiangxiang Zeng.
\newblock Prediction of multi-relational drug–gene interaction via dynamic hypergraph contrastive learning.
\newblock \emph{Briefings in Bioinformatics}, 24\penalty0 (6):\penalty0 bbad371, 10 2023.

\bibitem[Wang et~al.(2024)Wang, Xiao, Shang, and Peng]{dd_2}
Jingru Wang, Yihang Xiao, Xuequn Shang, and Jiajie Peng.
\newblock Predicting drug–target binding affinity with cross-scale graph contrastive learning.
\newblock \emph{Briefings in Bioinformatics}, 25\penalty0 (1):\penalty0 bbad516, 01 2024.

\bibitem[Lee et~al.(2023)Lee, Kim, Hyun, Lee, Kim, and Park]{gen_1}
Junseok Lee, Sungwon Kim, Dongmin Hyun, Namkyeong Lee, Yejin Kim, and Chanyoung Park.
\newblock Deep single-cell rna-seq data clustering with graph prototypical contrastive learning.
\newblock \emph{Bioinformatics}, 39\penalty0 (6):\penalty0 btad342, 05 2023.

\bibitem[Xiong et~al.(2023)Xiong, Luo, Shi, Liu, Xu, and Wang]{gen_2}
Zehao Xiong, Jiawei Luo, Wanwan Shi, Ying Liu, Zhongyuan Xu, and Bo~Wang.
\newblock scgcl: an imputation method for scrna-seq data based on graph contrastive learning.
\newblock \emph{Bioinformatics}, 39\penalty0 (3):\penalty0 btad098, 02 2023.

\bibitem[Ma et~al.(2022)Ma, Hu, Ge, Chen, Zhang, and Zhang]{social_1}
Guanghui Ma, Chunming Hu, Ling Ge, Junfan Chen, Hong Zhang, and Richong Zhang.
\newblock Towards robust false information detection on social networks with contrastive learning.
\newblock In \emph{Proceedings of the 31st ACM International Conference on Information \& Knowledge Management}, CIKM '22, page 1441–1450, New York, NY, USA, 2022. Association for Computing Machinery.
\newblock ISBN 9781450392365.
\newblock \doi{10.1145/3511808.3557477}.
\newblock URL \url{https://doi.org/10.1145/3511808.3557477}.

\bibitem[Sun et~al.(2022)Sun, Qian, Dong, Li, and Zhu]{social_2}
Tiening Sun, Zhong Qian, Sujun Dong, Peifeng Li, and Qiaoming Zhu.
\newblock Rumor detection on social media with graph adversarial contrastive learning.
\newblock In \emph{Proceedings of the ACM Web Conference 2022}, WWW '22, page 2789–2797, New York, NY, USA, 2022. Association for Computing Machinery.
\newblock ISBN 9781450390965.
\newblock \doi{10.1145/3485447.3511999}.
\newblock URL \url{https://doi.org/10.1145/3485447.3511999}.

\bibitem[Errica et~al.(2020)Errica, Podda, Bacciu, and Micheli]{errica_2020}
Federico Errica, Marco Podda, Davide Bacciu, and Alessio Micheli.
\newblock A fair comparison of graph neural networks for graph classification.
\newblock In \emph{International Conference on Learning Representations}, 2020.

\bibitem[Morris et~al.(2020)Morris, Kriege, Bause, Kersting, Mutzel, and Neumann]{tu_datasets}
Christopher Morris, Nils~M. Kriege, Franka Bause, Kristian Kersting, Petra Mutzel, and Marion Neumann.
\newblock Tudataset: A collection of benchmark datasets for learning with graphs.
\newblock In \emph{ICML 2020 Workshop on Graph Representation Learning and Beyond (GRL+ 2020)}, 2020.
\newblock URL \url{www.graphlearning.io}.

\bibitem[Chen et~al.(2024)Chen, Frias, and Gel]{topoGCL}
Yuzhou Chen, Jose Frias, and Yulia~R. Gel.
\newblock Topogcl: Topological graph contrastive learning.
\newblock \emph{Proceedings of the AAAI Conference on Artificial Intelligence}, 38\penalty0 (10):\penalty0 11453--11461, Mar. 2024.
\newblock \doi{10.1609/aaai.v38i10.29026}.
\newblock URL \url{https://ojs.aaai.org/index.php/AAAI/article/view/29026}.

\bibitem[Liu et~al.(2024)Liu, Wang, and Wu]{reinforce24}
Ziyang Liu, Chaokun Wang, and Cheng Wu.
\newblock Graph contrastive learning with reinforcement augmentation.
\newblock In Kate Larson, editor, \emph{Proceedings of the Thirty-Third International Joint Conference on Artificial Intelligence, {IJCAI-24}}, pages 2225--2233. International Joint Conferences on Artificial Intelligence Organization, 8 2024.
\newblock \doi{10.24963/ijcai.2024/246}.
\newblock URL \url{https://doi.org/10.24963/ijcai.2024/246}.
\newblock Main Track.

\bibitem[Liu et~al.(2023)Liu, Zhao, Wang, Geng, and Xiao]{MSSGCL}
Yanbei Liu, Yu~Zhao, Xiao Wang, Lei Geng, and Zhitao Xiao.
\newblock Multi-scale subgraph contrastive learning.
\newblock In Edith Elkind, editor, \emph{Proceedings of the Thirty-Second International Joint Conference on Artificial Intelligence, {IJCAI-23}}, pages 2215--2223. International Joint Conferences on Artificial Intelligence Organization, 8 2023.
\newblock \doi{10.24963/ijcai.2023/246}.
\newblock URL \url{https://doi.org/10.24963/ijcai.2023/246}.
\newblock Main Track.

\bibitem[Wu et~al.(2024)Wu, Wang, Han, and Ye]{CTaug}
Yucheng Wu, Leye Wang, Xiao Han, and Han-Jia Ye.
\newblock Graph contrastive learning with cohesive subgraph awareness.
\newblock In \emph{Proceedings of the ACM Web Conference 2024}, WWW '24, page 629–640, New York, NY, USA, 2024. Association for Computing Machinery.
\newblock ISBN 9798400701719.
\newblock \doi{10.1145/3589334.3645470}.
\newblock URL \url{https://doi.org/10.1145/3589334.3645470}.

\bibitem[Wang et~al.(2025)Wang, Xu, Geng, and Li]{khanGCL}
Zihu Wang, Boxun Xu, Hejia Geng, and Peng Li.
\newblock Khan-gcl: Kolmogorov-arnold network based graph contrastive learning with hard negatives, 2025.
\newblock URL \url{https://arxiv.org/abs/2505.15103}.

\bibitem[Hu et~al.(2020)Hu, Fey, Zitnik, Dong, Ren, Liu, Catasta, and Leskovec]{ogb}
Weihua Hu, Matthias Fey, Marinka Zitnik, Yuxiao Dong, Hongyu Ren, Bowen Liu, Michele Catasta, and Jure Leskovec.
\newblock Open graph benchmark: Datasets for machine learning on graphs.
\newblock In \emph{Advances in Neural Information Processing Systems}, 2020.

\bibitem[Wu et~al.(2018)Wu, Ramsundar, Feinberg, Gomes, Geniesse, Pappu, Leswing, and Pande]{ogb_molhiv_paper}
Zhenqin Wu, Bharath Ramsundar, Evan N. Feinberg, Joseph Gomes, Caleb Geniesse, Aneesh~S. Pappu, Karl Leswing, and Vijay Pande.
\newblock Moleculenet: a benchmark for molecular machine learning.
\newblock \emph{Chem. Sci.}, 9:\penalty0 513--530, 2018.
\newblock \doi{10.1039/C7SC02664A}.

\bibitem[Trivedi et~al.(2022{\natexlab{a}})Trivedi, Lubana, Heimann, Koutra, and Thiagarajan]{trivedi_two}
Puja Trivedi, Ekdeep~Singh Lubana, Mark Heimann, Danai Koutra, and Jayaraman~J. Thiagarajan.
\newblock Analyzing data-centric properties for graph contrastive learning.
\newblock In \emph{Advances in Neural Information Processing Systems}, 2022{\natexlab{a}}.

\bibitem[Trivedi et~al.(2022{\natexlab{b}})Trivedi, Lubana, Yan, Yang, and Koutra]{trivedi_one}
Puja Trivedi, Ekdeep~Singh Lubana, Yujun Yan, Yaoqing Yang, and Danai Koutra.
\newblock Augmentations in graph contrastive learning: Current methodological flaws \& towards better practices.
\newblock In \emph{Proceedings of the ACM Web Conference 2022}, WWW '22, page 1538–1549, New York, NY, USA, 2022{\natexlab{b}}. Association for Computing Machinery.
\newblock ISBN 9781450390965.
\newblock \doi{10.1145/3485447.3512200}.

\bibitem[Cai and Wang(2022)]{heuristic_cai_22}
Chen Cai and Yusu Wang.
\newblock A simple yet effective baseline for non-attributed graph classification, 2022.
\newblock URL \url{https://arxiv.org/abs/1811.03508}.

\bibitem[Gurukar et~al.(2022)Gurukar, Vijayan, srinivasan parthasarathy, Ravindran, Srinivasan, Bajaj, Cai, Keymanesh, Kumar, Maneriker, Mitra, and Patel]{gurukar2022benchmarking}
Saket Gurukar, Priyesh Vijayan, srinivasan parthasarathy, Balaraman Ravindran, Aakash Srinivasan, Goonmeet Bajaj, Chen Cai, Moniba Keymanesh, Saravana Kumar, Pranav Maneriker, Anasua Mitra, and Vedang Patel.
\newblock Benchmarking and analyzing unsupervised network representation learning and the illusion of progress.
\newblock \emph{Transactions on Machine Learning Research}, 2022.
\newblock ISSN 2835-8856.
\newblock URL \url{https://openreview.net/forum?id=GvF9ktXI1V}.

\bibitem[You et~al.(2020)You, Chen, Sui, Chen, Wang, and Shen]{gclyu}
Yuning You, Tianlong Chen, Yongduo Sui, Ting Chen, Zhangyang Wang, and Yang Shen.
\newblock Graph contrastive learning with augmentations.
\newblock In \emph{Advances in Neural Information Processing Systems}, 2020.

\bibitem[You et~al.(2021)You, Chen, Shen, and Wang]{joao}
Yuning You, Tianlong Chen, Yang Shen, and Zhangyang Wang.
\newblock Graph contrastive learning automated.
\newblock In Marina Meila and Tong Zhang, editors, \emph{Proceedings of the 38th International Conference on Machine Learning}, volume 139 of \emph{Proceedings of Machine Learning Research}, pages 12121--12132. PMLR, 18--24 Jul 2021.

\bibitem[Xia et~al.(2022)Xia, Wu, Chen, Hu, and Li]{simGRACE22}
Jun Xia, Lirong Wu, Jintao Chen, Bozhen Hu, and Stan~Z. Li.
\newblock Simgrace: A simple framework for graph contrastive learning without data augmentation.
\newblock In \emph{Proceedings of the ACM Web Conference 2022}, WWW '22, page 1070–1079, New York, NY, USA, 2022. Association for Computing Machinery.
\newblock ISBN 9781450390965.
\newblock \doi{10.1145/3485447.3512156}.

\bibitem[Sun et~al.(2020)Sun, Hoffman, Verma, and Tang]{Sun2020InfoGraph}
Fan-Yun Sun, Jordan Hoffman, Vikas Verma, and Jian Tang.
\newblock Infograph: Unsupervised and semi-supervised graph-level representation learning via mutual information maximization.
\newblock In \emph{International Conference on Learning Representations}, 2020.

\bibitem[Xu et~al.(2019)Xu, Hu, Leskovec, and Jegelka]{xu2018how}
Keyulu Xu, Weihua Hu, Jure Leskovec, and Stefanie Jegelka.
\newblock How powerful are graph neural networks?
\newblock In \emph{International Conference on Learning Representations}, 2019.

\bibitem[Li et~al.(2024)Li, Cao, Shuai, Miao, and Hwang]{rethinking_graphclass_li}
Zhengdao Li, Yong Cao, Kefan Shuai, Yiming Miao, and Kai Hwang.
\newblock Rethinking the effectiveness of graph classification datasets in benchmarks for assessing gnns.
\newblock In Kate Larson, editor, \emph{Proceedings of the Thirty-Third International Joint Conference on Artificial Intelligence, {IJCAI-24}}, pages 2144--2152. International Joint Conferences on Artificial Intelligence Organization, 8 2024.
\newblock \doi{10.24963/ijcai.2024/237}.
\newblock URL \url{https://doi.org/10.24963/ijcai.2024/237}.
\newblock Main Track.

\bibitem[Zhu et~al.(2021)Zhu, Xu, Liu, and Wu]{PyGCL21}
Yanqiao Zhu, Yichen Xu, Qiang Liu, and Shu Wu.
\newblock An empirical study of graph contrastive learning.
\newblock In \emph{Thirty-fifth Conference on Neural Information Processing Systems Datasets and Benchmarks Track (Round 2)}, 2021.

\bibitem[Hu* et~al.(2020)Hu*, Liu*, Gomes, Zitnik, Liang, Pande, and Leskovec]{GINE}
Weihua Hu*, Bowen Liu*, Joseph Gomes, Marinka Zitnik, Percy Liang, Vijay Pande, and Jure Leskovec.
\newblock Strategies for pre-training graph neural networks.
\newblock In \emph{International Conference on Learning Representations}, 2020.
\newblock URL \url{https://openreview.net/forum?id=HJlWWJSFDH}.

\bibitem[Guo et~al.(2023)Guo, Wang, Wei, and Wang]{23-architecture_matters}
Xiaojun Guo, Yifei Wang, Zeming Wei, and Yisen Wang.
\newblock Architecture matters: Uncovering implicit mechanisms in graph contrastive learning.
\newblock In \emph{Advances in Neural Information Processing Systems}, 2023.
\newblock URL \url{https://proceedings.neurips.cc/paper_files/paper/2023/file/5acf5a0ee5c17d372bfe7fdaeffd6e33-Paper-Conference.pdf}.

\bibitem[Fey and Lenssen(2019)]{pyg1}
Matthias Fey and Jan~E. Lenssen.
\newblock Fast graph representation learning with {PyTorch Geometric}.
\newblock In \emph{ICLR Workshop on Representation Learning on Graphs and Manifolds}, 2019.

\bibitem[Fey et~al.(2025)Fey, Sunil, Nitta, Puri, Shah, Stojanovi{\'c}, Bendias, Alexandria, Kocijan, Zhang, He, Lenssen, and Leskovec]{pyg2}
Matthias Fey, Jinu Sunil, Akihiro Nitta, Rishi Puri, Manan Shah, Bla{\v{z}} Stojanovi{\'c}, Ramona Bendias, Barghi Alexandria, Vid Kocijan, Zecheng Zhang, Xinwei He, Jan~E. Lenssen, and Jure Leskovec.
\newblock {PyG} 2.0: Scalable learning on real world graphs.
\newblock In \emph{Temporal Graph Learning Workshop @ KDD}, 2025.

\bibitem[van~den Oord et~al.(2019)van~den Oord, Li, and Vinyals]{oord2019representationlearningcontrastivepredictive}
Aaron van~den Oord, Yazhe Li, and Oriol Vinyals.
\newblock Representation learning with contrastive predictive coding, 2019.
\newblock URL \url{https://arxiv.org/abs/1807.03748}.

\bibitem[Shervashidze et~al.(2011)Shervashidze, Schweitzer, van Leeuwen, Mehlhorn, and Borgwardt]{WL-kerenl}
Nino Shervashidze, Pascal Schweitzer, Erik~Jan van Leeuwen, Kurt Mehlhorn, and Karsten~M. Borgwardt.
\newblock Weisfeiler-lehman graph kernels.
\newblock \emph{Journal of Machine Learning Research}, 12\penalty0 (77):\penalty0 2539--2561, 2011.
\newblock URL \url{http://jmlr.org/papers/v12/shervashidze11a.html}.

\bibitem[Yanardag and Vishwanathan(2015)]{DGK}
Pinar Yanardag and S.V.N. Vishwanathan.
\newblock Deep graph kernels.
\newblock In \emph{Proceedings of the 21th ACM SIGKDD International Conference on Knowledge Discovery and Data Mining}, KDD '15, page 1365–1374, New York, NY, USA, 2015. Association for Computing Machinery.
\newblock ISBN 9781450336642.
\newblock \doi{10.1145/2783258.2783417}.
\newblock URL \url{https://doi.org/10.1145/2783258.2783417}.

\bibitem[Adhikari et~al.(2018)Adhikari, Zhang, Ramakrishnan, and Prakash]{sub2vec}
Bijaya Adhikari, Yao Zhang, Naren Ramakrishnan, and B.~Aditya Prakash.
\newblock Sub2vec: Feature learning for subgraphs.
\newblock In \emph{Advances in Knowledge Discovery and Data Mining: 22nd Pacific-Asia Conference, PAKDD 2018, Melbourne, VIC, Australia, June 3-6, 2018, Proceedings, Part II}, page 170–182, Berlin, Heidelberg, 2018. Springer-Verlag.
\newblock ISBN 978-3-319-93036-7.
\newblock \doi{10.1007/978-3-319-93037-4_14}.
\newblock URL \url{https://doi.org/10.1007/978-3-319-93037-4_14}.

\bibitem[Grover and Leskovec(2016)]{node2vec}
Aditya Grover and Jure Leskovec.
\newblock node2vec: Scalable feature learning for networks.
\newblock In \emph{Proceedings of the 22nd ACM SIGKDD International Conference on Knowledge Discovery and Data Mining}, KDD '16, page 855–864, New York, NY, USA, 2016. Association for Computing Machinery.
\newblock ISBN 9781450342322.
\newblock \doi{10.1145/2939672.2939754}.
\newblock URL \url{https://doi.org/10.1145/2939672.2939754}.

\bibitem[Narayanan et~al.(2017)Narayanan, Chandramohan, Venkatesan, Chen, Liu, and Jaiswal]{graph2vec}
Annamalai Narayanan, Mahinthan Chandramohan, Rajasekar Venkatesan, Lihui Chen, Yang Liu, and Shantanu Jaiswal.
\newblock graph2vec: Learning distributed representations of graphs, 2017.
\newblock URL \url{https://arxiv.org/abs/1707.05005}.

\bibitem[Kingma and Ba(2015)]{Kingma2015AdamAM}
Diederik~P. Kingma and Jimmy Ba.
\newblock Adam: A method for stochastic optimization.
\newblock In \emph{ICLR}, 2015.

\end{thebibliography}

\newpage
\appendix
\section{Expanded Related Work} \label{rel_work}

\textbf{Graph contrastive learning.}
GCL learns node- or graph-level representations by contrasting positive and negative views generated from the data itself \cite{ju2024gclreview}. We focus on methods for \emph{graph classification}. See App.~\ref{background} for a brief primer and the specific methods considered.

\textbf{Benchmarking practices.}
There is a sustained effort to improve evaluation in graph learning. Errica et al.\ \cite{errica_2020} advocated fixed splits, multiple seeds, and simple baselines (e.g., MoleculeFingerprint+MLP) for supervised graph classification. Gurukar et al.\ \cite{gurukar2022benchmarking} examined unsupervised settings and introduced heuristics for link prediction and node classification, motivating our handcrafted statistics baseline. Li et al.\ \cite{rethinking_graphclass_li} questioned the suitability of TU datasets, showing that simple graph properties can correlate strongly with labels. To the best of our knowledge, few works systematically compare \emph{GCL} against untrained or handcrafted baselines on graph classification. Our study aims to fill that gap.

\textbf{GCL-specific evaluations.}
Trivedi et al.\ showed that domain-agnostic augmentations can erase task signal on small TU datasets and recommended sanity checks such as untrained baselines \cite{trivedi_one}. Guo et al.\ found that several GCL behaviors arise from encoder design and normalization, with weaker sensitivity to augmentations than in vision CL \cite{23-architecture_matters}. Trivedi et al.\ also analyzed invariance and recoverability for graph-level GCL, reported marginal gains over untrained encoders on standard benchmarks, and released the synthetic generator we adopt \cite{trivedi_two}. Our work complements these studies by adding the under-explored perspective of \emph{dataset scale} when comparing GCL with untrained and simple baselines.

\section{Background on Graph Contrastive Learning} \label{background}

In this work we study the following widely cited GCL methods: GraphCL, InfoGraph, JOAO and SimGRACE \cite{gclyu, Sun2020InfoGraph, joao, simGRACE22}. The first two form the core of our scaling and synthetic-dataset analysis, while JOAO and SimGRACE are included in our TU-benchmark comparisons for completeness. These methods are widely cited and remain standard reference points in the literature for graph-level GCL. All methods are implemented using \texttt{PyTorch Geometric} \cite{pyg1, pyg2} and the \texttt{PyGCL} library \cite{PyGCL21}.

\textbf{GraphCL.} GraphCL \cite{gclyu} generates two correlated views of a graph by applying random augmentations, and trains an encoder to maximize agreement between these views using the InfoNCE loss \cite{oord2019representationlearningcontrastivepredictive}. Given an input graph $G$, a GNN encoder $f_\theta$ extracts a graph-level vector $h$. This vector is then passed through a projection head $p_\phi$ (a simple MLP) to obtain the final representation $z = p_\phi(h)$. For each minibatch of $N$ graphs, two augmentations are applied to every input, producing $2N$ projected embeddings $\{z_{n,i}\}$ where $i \in \{1,2\}$ indexes the augmented view. 

The contrastive objective encourages representations of two augmented views of the same graph (positives) to be similar while pushing apart embeddings of different graphs (negatives). For the $n$-th graph, the cross-entropy loss is:
\begin{equation}
\ell_n = -\log 
\frac{\exp(\mathrm{sim}(z_{n,1}, z_{n,2})/\tau)}
{\sum_{n'=1,\, n'\neq n}^N \exp(\mathrm{sim}(z_{n,1}, z_{n',2})/\tau)},
\end{equation}
where $\mathrm{sim}(\cdot,\cdot)$ denotes cosine similarity and $\tau$ is a temperature parameter. For brevity we show one direction; in practice we average both directions over the two augmented views. The final loss averages $\ell_n$ across all graphs in the minibatch. GraphCL remains a popular baseline in the GCL literature due to its simplicity and strong empirical performance. Following prior work, we focus on two standard augmentation strategies: \emph{node-dropping} and \emph{edge-dropping}, as well as their combination (node+edge). For both, nodes and edges are stochastically removed with probability $p_r$, which is a hyperparameter to be chosen by the user \cite{PyGCL21}.

\textbf{InfoGraph.} InfoGraph \cite{Sun2020InfoGraph} instead maximizes mutual information between local and global representations within the same graph. Specifically, embeddings of substructures (nodes or patches) are encouraged to align with the graph-level embedding of their parent graph, while embeddings from different graphs act as negatives. Unlike GraphCL, InfoGraph does not rely on stochastic augmentations: it operates directly on the input graph, using a discriminator to score alignment between global and local features. This makes it less sensitive to augmentation choices and, as we later observe, more robust on synthetic datasets where augmentations can obscure the task-defining motif signal.

\textbf{JOAO.} JOAO \cite{joao} builds on GraphCL by automatically selecting augmentation strategies during training via a policy optimization scheme, rather than fixing them a priori. We include it in our TU-benchmark study as a representative variant of augmentation-search methods.

\textbf{SimGRACE.} SimGRACE \cite{simGRACE22} is a lightweight alternative that avoids graph augmentations altogether, instead injecting noise directly into the model parameters to create different ``views'' of the encoder. We include it for completeness in TU-dataset comparisons.

\section{Benchmarking TU Datasets} \label{more_TU}
\subsection{Evaluation Protocol and Dataset Statistics} \label{eval_protocol}
Early GCL papers popularized an unsupervised \emph{embedding probe} to assess representation quality: pretrain a GNN contrastively on the full dataset, freeze the encoder, extract graph embeddings, and fit a linear SVM with 10-fold stratified cross-validation \cite{Sun2020InfoGraph}. We report graph-classification accuracy averaged over 5 seeds (with C selected by cross-validation on the training folds). 

While other evaluations exist, the embedding probe remains central in graph-level GCL, and it is the protocol we study here. Many recent graph-level GCL papers \cite{topoGCL,reinforce24,MSSGCL,CTaug,khanGCL} still evaluate on the TU Dortmund graph-classification suite \cite{tu_datasets}. While the exact subset of datasets used varies slightly across papers, we fix our analysis to the datasets in Table~\ref{tab:dataset-stats}.

\begin{table}[h]
\centering
\caption{Summary statistics of TU‑Dortmund graph classification datasets used in our work.}
\label{tab:dataset-stats}
\begin{tabular}{l l r r r r}
\toprule
Dataset         & Domain        & \#Graphs & \#Classes & Avg.\,\#Nodes & Avg.\,\#Edges \\
\midrule
MUTAG           & Biological    &   188    &     2     &    17.93      &     19.79     \\
PROTEINS        & Biological    &  1113    &     2     &    39.06      &     72.82     \\
NCI1            & Biological    &  4110    &     2     &    29.87      &     32.30     \\
DD              & Biological    &  1178    &     2     &   284.32      &    715.66     \\
\midrule
COLLAB          & Social-net    &  5000    &     3     &    74.49      &   2457.76     \\
IMDB-BINARY     & Social-net    &  1000    &     2     &    19.77      &     96.53     \\
REDDIT-BINARY   & Social-net    &  2000    &     2     &   429.63      &    497.75     \\
\bottomrule
\end{tabular}
\end{table}

\subsection{Detailed benchmarking results: TU datasets} \label{benchmark_res}

 \textbf{Baseline Architectures.} For GCL methods we adopt the standard 3-layer GIN encoder used in InfoGraph~\cite{Sun2020InfoGraph}, with \texttt{hidden\_dim}=32 and sum pooling across nodes. Outputs from each layer are pooled and concatenated to form the final graph embedding. As in Errica et al.~\cite{errica_2020}, MolFingerprint is a two-layer MLP with \texttt{hidden\_dim}=256, applied to node features followed by global pooling. It ignores message passing and treats the graph as a bag of node attributes.

 Besides the baselines we test in Sec. \ref{benchmarking_simple}, we also include comparisons with graph kernels \cite{WL-kerenl,DGK} and shallow embedding techniques \cite{sub2vec, node2vec, graph2vec}. Results are quoted from previous papers \cite{MSSGCL}, with matching experimental settings as ours. Table \ref{tab:super-comparison} is a compilation of all benchmarking results and table \ref{tab:untrained-vs-trivial} provides some ablations for the handcrafted statistics baseline.
 
\begin{table*}[h]
  \centering
  \scriptsize
  \setlength{\tabcolsep}{4pt}
  \caption{Comparison of unsupervised GCL methods, classical graph kernels, and simple baselines on the same 10-fold SVM protocol (mean \(\pm\) std) in Sec. \ref{eval_protocol}. MolFingerprint denotes the molecular fingerprint + MLP baseline, which is only evaluated on biological datasets. (N/A = not available - method requires node features of molecular graphs). Top 3 in each column are in \textbf{bold}}
  \label{tab:super-comparison}
  \begin{tabular}{lccccccc}
    \toprule
    Method            & NCI1           & PROTEINS       & DD             & MUTAG          & COLLAB         & RDT-B          & IMDB-B         \\
    \midrule
    \textbf{Graph kernels} &                    &                &                &                &                &                &                \\
    \quad WL kernel   & \textbf{80.01 \(\pm\) 0.50}   & 72.92 \(\pm\) 0.56   & 74.02 \(\pm\) 2.28   & 80.72 \(\pm\) 3.00   & 60.30 \(\pm\) 3.44   & 68.82 \(\pm\) 0.41   & \textbf{72.30 \(\pm\) 3.44}   \\
    \quad DGK         & \textbf{80.31 \(\pm\) 0.46}   & 73.30 \(\pm\) 0.82   & 74.85 \(\pm\) 0.74   & 87.44 \(\pm\) 2.72   & 64.66 \(\pm\) 0.50   & 78.04 \(\pm\) 0.39   & 66.96 \(\pm\) 0.56   \\
    \midrule
    \textbf{Shallow embeddings} &                &                    &                    &                    &                    &                    &                  \\
    \quad sub2vec     & 52.84 \(\pm\) 1.47   & 53.03 \(\pm\) 5.55   & 54.33 \(\pm\) 2.44   & 61.05 \(\pm\)15.80  & 55.26 \(\pm\) 1.54   & 71.48 \(\pm\) 0.41   & 55.26 \(\pm\) 1.54   \\
    \quad node2vec    & 54.89 \(\pm\) 1.61   & 57.49 \(\pm\) 3.57   & 74.77 \(\pm\) 0.51   & 72.63 \(\pm\)10.20  & 54.57 \(\pm\) 0.37   & 72.76 \(\pm\) 0.92   & 38.60 \(\pm\) 2.30   \\
    \quad graph2vec   & 73.22 \(\pm\) 1.81   & 73.30 \(\pm\) 2.05   & 70.32 \(\pm\) 2.32   & 83.15 \(\pm\) 9.25   & \textbf{71.10 \(\pm\) 0.54}   & 75.48 \(\pm\) 1.03   & 71.10 \(\pm\) 0.54   \\
    \midrule
    \textbf{GCL methods} &                    &                    &                    &                    &                    &                    &                  \\
    \quad InfoGraph   & 76.20 \(\pm\) 1.06   & 74.44 \(\pm\) 0.31   & 72.85 \(\pm\) 1.78   & \textbf{89.01 \(\pm\) 1.13}   & 70.65 \(\pm\) 1.13   & 82.50 \(\pm\) 1.42   & \textbf{73.03 \(\pm\) 0.87}   \\
    \quad GraphCL     & 77.87 \(\pm\) 0.41   & 74.39 \(\pm\) 0.45   & \textbf{78.62 \(\pm\) 0.40}   & 86.80 \(\pm\) 1.34   & \textbf{71.36 \(\pm\) 1.15}   & \textbf{89.53 \(\pm\) 0.84}   & 71.14 \(\pm\) 0.44   \\
    \quad JOAO        & 78.07 \(\pm\) 0.47   & 74.55 \(\pm\) 0.41   & 77.32 \(\pm\) 0.54   & 87.35 \(\pm\) 1.02   & 69.50 \(\pm\) 0.36   & 85.29 \(\pm\) 1.35   & 70.21 \(\pm\) 3.08   \\
    \quad JOAO V2     & 78.36 \(\pm\) 0.53   & 74.07 \(\pm\) 1.10   & 77.40 \(\pm\) 1.15   & 87.67 \(\pm\) 0.79   & 69.33 \(\pm\) 0.34   & 86.42 \(\pm\) 1.45   & 70.83 \(\pm\) 0.25   \\
    \quad SimGRACE    & \textbf{79.12 \(\pm\) 0.44}   & \textbf{75.35 \(\pm\) 0.09}   & \textbf{77.44 \(\pm\)} 1.11   & \textbf{89.01 \(\pm\) 1.31}   & \textbf{71.72 \(\pm\) 0.82}   & \textbf{89.51 \(\pm\) 0.89}   & 71.30 \(\pm\) 0.77   \\
    \midrule
    \textbf{Simple baselines} &                &                    &                    &                    &                    &                    &                  \\
    \quad Untrained GNN & 72.73 \(\pm\) 0.86 & \textbf{74.61 \(\pm\) 0.65} & 77.30 \(\pm\) 0.44 & \textbf{88.26 \(\pm\) 0.57} & 62.88 \(\pm\) 0.05 & 76.86 \(\pm\) 0.31 & 68.20 \(\pm\) 0.22 \\
    \quad Handcrafted statistics   & 68.50 \(\pm\) 0.47 & 73.80 \(\pm\) 0.31 & 76.25 \(\pm\) 0.33 & 86.19 \(\pm\) 1.33 & 69.46 \(\pm\) 0.30 & \textbf{87.72 \(\pm\) 0.22} & 70.06 \(\pm\) 0.45 \\
    \quad MolFingerprint & 69.43 \(\pm\) 0.10 & \textbf{75.71 \(\pm\) 0.38} & \textbf{78.59 \(\pm\) 0.48} & 84.05 \(\pm\) 0.35 & N/A              & N/A              & N/A               \\
    \bottomrule
  \end{tabular}
\end{table*}

\begin{table*}[h]
  \centering
  \scriptsize
  \setlength{\tabcolsep}{4pt}
  \caption{Ablations of the handcrafted graph statistics baseline (mean \(\pm\) std) on TU datasets. We isolate each of the individual statistics components to see which one performs the best. We also include a random baseline which is a random gaussian vector of the same dimension.}
  \label{tab:untrained-vs-trivial}
  \begin{tabular}{lccccccc}
    \toprule
    Method                      & NCI1             & PROTEINS         & DD               & MUTAG            & COLLAB           & RDT‑B            & IMDB‑B           \\
    \midrule

    \textit{Trivial statistics:}&                   &                    &                    &                    &                    &                    &                  \\
    \quad node count            & 62.65 \(\pm\) 0.18 & 71.81 \(\pm\) 0.37 & 76.05 \(\pm\) 0.15 & 83.85 \(\pm\) 0.93 & 56.22 \(\pm\) 0.02 & 76.61 \(\pm\) 0.05 & 53.86 \(\pm\) 0.62 \\
    \quad avg.\ degree          & 55.91 \(\pm\) 0.20 & 60.72 \(\pm\) 0.72 & 63.48 \(\pm\) 0.17 & 85.36 \(\pm\) 0.44 & 69.05 \(\pm\) 0.05 & 61.87 \(\pm\) 0.32 & 71.38 \(\pm\) 0.19 \\
    \quad deg.\ histogram       & 68.24 \(\pm\) 0.36 & 73.80 \(\pm\) 0.38 & 75.94 \(\pm\) 0.33 & 87.15 \(\pm\) 0.22 & 58.89 \(\pm\) 0.15 & 80.56 \(\pm\) 0.24 & 62.78 \(\pm\) 0.43 \\
    \quad all trivial (1+2+3)   & 68.50 \(\pm\) 0.47 & 73.80 \(\pm\) 0.31 & 76.25 \(\pm\) 0.33 & 86.19 \(\pm\) 1.33 & 69.46 \(\pm\) 0.30 & 87.72 \(\pm\) 0.22 & 70.06 \(\pm\) 0.45 \\
    \midrule
    Random baseline & 49.18 \(\pm\) 1.06 & 59.51 \(\pm\) 0.07 & 58.44 \(\pm\) 0.44 & 65.95 \(\pm\) 0.58 & 52.01 \(\pm\) 0.01 & 49.63 \(\pm\) 1.58 & 50.98 \(\pm\) 2.01 \\
    \bottomrule
  \end{tabular}
\end{table*}

\newpage
\section{Experimental Configurations and Details: \textsc{ogbg-molhiv}} \label{hpo}

\textsc{ogbg-molhiv} is a molecular graph dataset from Open Graph Benchmark (OGB) collection. It contains 41,127 graphs, with 9-dimensional node features and 3-dimensional edge features \cite{ogb, ogb_molhiv_paper}. The task is to predict HIV inhibition (binary). We use OGB's official scaffold splits and ROC-AUC evaluation guidelines.

\textbf{GINE encoder (MolHIV).} We use a 3-layer edge-aware GINE encoder with \texttt{hidden\_dim}=32 and batch norm after each layer. For graph embeddings, we apply global add pooling at every layer and concatenate the pooled outputs, yielding a 96-d vector. The projection head dimension is set to \texttt{proj\_dim}=32 for contrastive pretraining.

\textbf{Evaluation protocol.} We follow the embedding–probe setup for \textsc{ogbg-molhiv} proposed by \cite{PyGCL21}. Encoders are frozen and graph embeddings are extracted, after which a logistic regression classifier is trained to predict labels. We implement the probe using \texttt{scikit-learn}’s \texttt{LogisticRegression} with class weights balanced to account for label skew, and scale inputs with a \texttt{StandardScaler}. The regularization strength $C$ is tuned over $\{0.01, 0.1, 1, 10\}$ on the validation split. For evaluation we use the official OGB \texttt{Evaluator}, reporting ROC-AUC on the held-out test set.

\textbf{Hyperparameter search and final settings.} 
Since GraphCL relies on augmentations, we conducted a small grid search over three augmentation strategies: node-dropping (ND), edge-dropping (ED), and their combination (ND+ED). The drop probability $p_r$ was varied over $\{0.05, 0.1\}$, the hidden encoder dimension over $\{32, 64\}$, and the learning rate over $\{10^{-2}, 10^{-3}\}$. Experiments were run with 5 random seeds on a 50\% fraction of \textsc{ogbg-molhiv} for 100 epochs. 


The final configuration was trained with the Adam optimizer \cite{Kingma2015AdamAM} with InfoNCE loss \cite{oord2019representationlearningcontrastivepredictive} and the following hyperparameters:  

\begin{table}[h]
\centering
\caption{Final GraphCL configuration on \textsc{ogbg-molhiv}. \emph{Abbreviations:} \texttt{LR} learning rate, \texttt{WD} weight decay, \texttt{TEMP} temperature, \texttt{PROJ} projection dimension, \texttt{HIDDEN} hidden dimension (GNN).}
\label{tab:molhiv-hparams}
\begin{tabular}{l l l l l l l}
\toprule
\texttt{EPOCHS} & \texttt{BATCH} & \texttt{LR} & \texttt{WD} & \texttt{TEMP} & \texttt{PROJ} & \texttt{HIDDEN} \\
\midrule
100 & 256 & \(10^{-3}\) & \(10^{-5}\) & 0.2 & 64 & 32 \\
\midrule
\texttt{AUG} & \multicolumn{6}{l}{Edge drop (\(p=0.10\))} \\
\bottomrule
\end{tabular}
\end{table}

\section{Experimental Configurations and Details: Synthetic Dataset} \label{synth_info}
Trivedi et al.’s generator \cite{trivedi_two} constructs synthetic graph-classification datasets with controllable difficulty. Each graph consists of a label-defining \emph{motif} graph (which they refer to as \emph{content}) attached to a label-agnostic random \emph{background} tree graph (referred to as \emph{style}). In their original paper, the style multiplier $S$ scales the background and thereby controls task difficulty, while the total dataset size is varied independently. While they used this generator to design augmentations that preserve motif signal, we instead use it as a tool to probe how GCL behaves at scale, since it provides a simple way to increase both task difficulty and dataset size. In our experiments, we instantiate the six distinct motif graphs from their work, sweep $S \in {2,4,6}$, and vary dataset size over ${600, 3{,}000, 6{,}000, 12{,}000}$ graphs. We also assign a fixed 10-dimensional node feature vector to both style and content nodes. The same backbone GIN architecture and evaluation protocol described in App.~\ref{more_TU} are used.

\textbf{Training setup (synthetic datasets).}
Both GraphCL and InfoGraph were trained with fixed settings across all dataset sizes and difficulties (using the Adam optimizer). Shared parameters were:
\texttt{EPOCHS}=50, \texttt{BATCH}=256, \texttt{HIDDEN}=32 (GNN hidden dimension), and \texttt{LAYERS}=3 (GIN layers).

\begin{table}[h]
\centering
\caption{Training configuration for synthetic datasets. 
\emph{Abbreviations:} \texttt{LR} = learning rate, \texttt{PROJ} = projection dimension, 
\texttt{P} = drop probability}
\label{tab:synth-hparams}
\begin{tabular}{l l l l l l}
\toprule
Method & \texttt{EPOCHS} & \texttt{BATCH} & \texttt{HIDDEN}/\texttt{LAYERS} & \texttt{LR} & Other settings \\
\midrule
GraphCL & 50 & 256 & 32 / 3 & $10^{-3}$ & \texttt{PROJ}=32, ND+ED, $p=0.15$ \\
InfoGraph & 50 & 256 & 32 / 3 & $10^{-3}$ & \texttt{PROJ}=2 \\
\bottomrule
\end{tabular}
\end{table}

\textbf{Supplementary figures.}

\begin{figure}[h]
  \centering

  \begin{subfigure}{0.7\linewidth}
    \centering 
    \includegraphics[width=\linewidth]{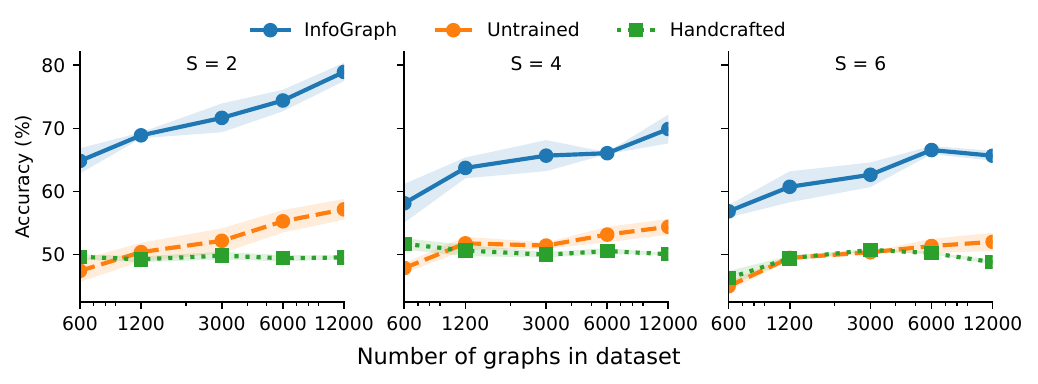}
  \end{subfigure}

  \begin{subfigure}{0.7\linewidth}
    \centering
    \includegraphics[width=\linewidth]{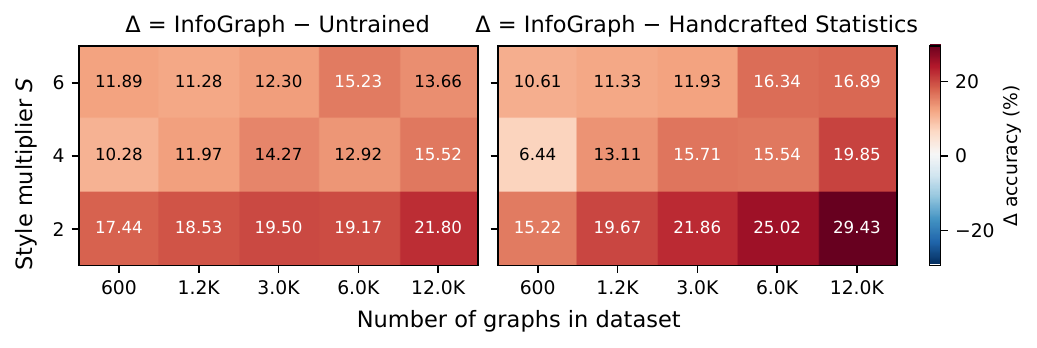}
  \end{subfigure}

  \caption{\textbf{Scaling on Synthetic Datasets.}
    \emph{Top:} InfoGraph. Test accuracy vs.\ total number of graphs (log scale) for $S\in\{2,4,6\}$; bands show $\pm1\sigma$ over seeds; handcrafted baseline shown as dotted line.
  \emph{Bottom:} $\Delta$ heatmaps for InfoGraph — Untrained and InfoGraph — Handcrafted across style multiplier $S$ and dataset size (shared color scale; positive values favor InfoGraph).}
  \label{fig:infograph_composite}
\end{figure}

\section{Reproducibility Statement}
We will release all code upon acceptance. All models are trained on an NVIDIA Quadro RTX 6000 (24 GB).

\end{document}